\documentclass[conference]{IEEEtran}
\IEEEoverridecommandlockouts
\usepackage{cite}
\usepackage{amsmath,amssymb,amsfonts, amsthm}
\usepackage{algorithmic}
\usepackage{graphicx}
\usepackage{textcomp}
\usepackage{xcolor}
\usepackage{subfigure}
\def\BibTeX{{\rm B\kern-.05em{\sc i\kern-.025em b}\kern-.08em?
    T\kern-.1667em\lower.7ex\hbox{E}\kern-.125emX}}

\newcommand{\vlat}{\vz}
\newcommand{\lat}{z}
\newcommand{\vnat}{\vlambda}
\newcommand{\vmean}{\vmu}
\newcommand{\nat}{\lambda}
\newcommand{\mean}{\mu}
\newcommand{\fim}{\vF}
\newcommand{\vgradmean}{\tilde{\vg}}
\newcommand{\gradmean}{\tilde{g}}
\newcommand{\vpriorpar}{\veta_0}

\newcommand\cut[1]{}

\newcommand{\elbofinal}{\mathcal{L}}

\newcommand{\squishlist}{
   \begin{list}{$\bullet$}
    { \setlength{\itemsep}{0pt}      \setlength{\parsep}{3pt}
      \setlength{\topsep}{3pt}       \setlength{\partopsep}{0pt}
      \setlength{\leftmargin}{1.5em} \setlength{\labelwidth}{1em}
      \setlength{\labelsep}{0.5em} } }

\newcommand{\squishlisttwo}{
   \begin{list}{$\bullet$}
    { \setlength{\itemsep}{0pt}    \setlength{\parsep}{0pt}
      \setlength{\topsep}{0pt}     \setlength{\partopsep}{0pt}
      \setlength{\leftmargin}{2em} \setlength{\labelwidth}{1.5em}
      \setlength{\labelsep}{0.5em} } }

\newcommand{\squishend}{
    \end{list}  }

{}
\newtheorem{thm}{Theorem}{}
{}
{}
{}
{}

\newenvironment{myproof}{{\bf Proof: }}{}
\newenvironment{example}{\emph{Example: }}{}

\newcommand{\half}{\mbox{$\frac{1}{2}$}}

\newcommand{\real}{\mbox{$\mathbb{R}$}}

\newcommand{\sqr}[1]{\left[#1\right]}
\newcommand{\crl}[1]{\left\{#1\right\}}

\newcommand{\myexpect}{\mathbb{E}}

\newcommand{\gauss}{\mbox{${\cal N}$}}

\newcommand{\myvec}[1]{\mbox{$\mathbf{#1}$}}
\newcommand{\myvecsym}[1]{\mbox{$\boldsymbol{#1}$}}

\newcommand{\veta}{\mbox{$\myvecsym{\eta}$}}

\newcommand{\vmu}{\mbox{$\myvecsym{\mu}$}}
\newcommand{\vlambda}{\mbox{$\myvecsym{\lambda}$}}

\newcommand{\vphi}{\mbox{$\myvecsym{\phi}$}}

\newcommand{\vg}{\mbox{$\myvec{g}$}}

\newcommand{\vm}{\mbox{$\myvec{m}$}}

\newcommand{\vx}{\mbox{$\myvec{x}$}}

\newcommand{\vz}{\mbox{$\myvec{z}$}}

\newcommand{\vF}{\mbox{$\myvec{F}$}}

\newcommand{\vH}{\mbox{$\myvec{H}$}}
\newcommand{\vI}{\mbox{$\myvec{I}$}}

\newcommand{\vV}{\mbox{$\myvec{V}$}}

\newcommand{\calD}{\mbox{${\cal D}$}}

\newcommand{\data}{\calD}

\newcommand{\be}{\begin{equation}}
\newcommand{\ee}{\end{equation}}
\newcommand{\bea}{\begin{eqnarray}}
\newcommand{\eea}{\end{eqnarray}}
\newcommand{\beaa}{\begin{eqnarray*}}
\newcommand{\eeaa}{\end{eqnarray*}}

\usepackage{todonotes}

\begin{document}

\title{Fast yet Simple Natural-Gradient Descent for Variational Inference in Complex Models}

\author{\IEEEauthorblockN{Mohammad Emtiyaz Khan}
\IEEEauthorblockA{\textit{RIKEN Center for Advanced Intelligence Project}\\
Tokyo, Japan\\
emtiyaz.khan@riken.jp} 
\and
\IEEEauthorblockN{Didrik Nielsen}
\IEEEauthorblockA{\textit{RIKEN Center for Advanced Intelligence Project}\\
Tokyo, Japan\\
didrik.nielsen@riken.jp}
}

\maketitle

\begin{abstract}
   Bayesian inference plays an important role in advancing machine learning, but faces computational challenges when applied to complex models such as deep neural networks.
Variational inference circumvents these challenges by formulating Bayesian inference as an optimization problem and solving it using gradient-based optimization.
In this paper, we argue in favor of \emph{natural-gradient} approaches which, unlike their \emph{gradient}-based counterparts, can improve convergence by exploiting the information geometry of the solutions.
We show how to derive fast yet simple natural-gradient updates by using a duality associated with exponential-family distributions. 
An attractive feature of these methods is that, by using natural-gradients, they are able to extract accurate local approximations for individual model components.
We summarize recent results for Bayesian deep learning showing the superiority of natural-gradient approaches over their gradient counterparts.

\end{abstract}


\section{Introduction}
\label{sec:intro}
Modern machine-learning methods, such as deep learning, are capable of producing accurate predictions which has lead to their enormous recent success in fields, e.g., computer vision, speech recognition, and recommendation systems.
However, this is not enough for other fields such as robotics and medical diagnostics where we also require an accurate estimate of confidence or uncertainty in the predictions.
Bayesian inference provides such uncertainty measures by using the \emph{posterior distribution} obtained using Bayes' rule. Unfortunately, this computation requires integrating over all possible values of the model parameters, which is infeasible for large complex models such as Bayesian neural networks.

Sampling methods such as Markov Chain Monte Carlo usually converge slowly when applied to such large problems. In contrast, approximate Bayesian methods such as variational inference (VI) can scale to large problems by obtaining approximations to the posterior distribution by using an optimization method, e.g., stochastic-gradient descent (SGD) methods \cite{graves2011practical, ranganath2013black, blundell2015weight}. These methods could provide reasonable approximations very quickly.

\begin{figure*}[t]
\centering
\subfigure[Gradient-based methods use the Euclidean distance which is a poor metric to measure distance between distributions. The bottom two distributions are almost identical while the top ones barely overlap, yet Euclidean distance is the same.]{\includegraphics[height=1.9in]{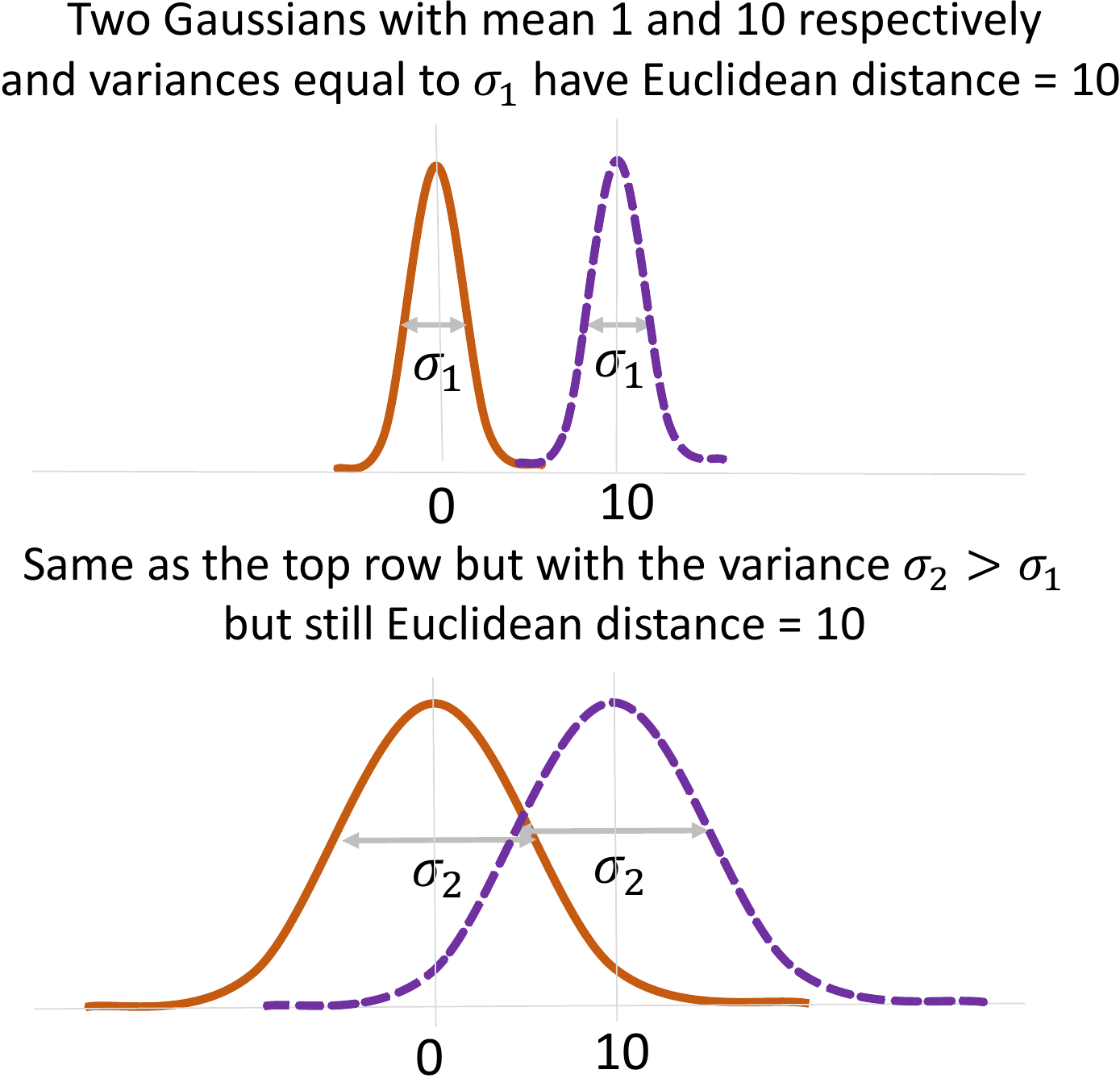} \label{figure:gaussians}}
\hfill
\subfigure[Natural-gradient method could converge faster than gradient-based methods. We apply a Bayesian neural network on two datasets, namely the Australian dataset (shown in left) and the Breast Cancer dataset (shown in right). A lower value of the test $\log_2$loss is considered better. The natural-gradient method is the Variational Online Gauss-Newton (VOGN) method proposed in \cite{vadam2018} while the gradient method is the Bayes-by-Backprop method proposed in
\cite{blundell2015weight}. The latter uses the Adam optimizer \cite{kingma2014adam}.]{\includegraphics[height=1.9in]{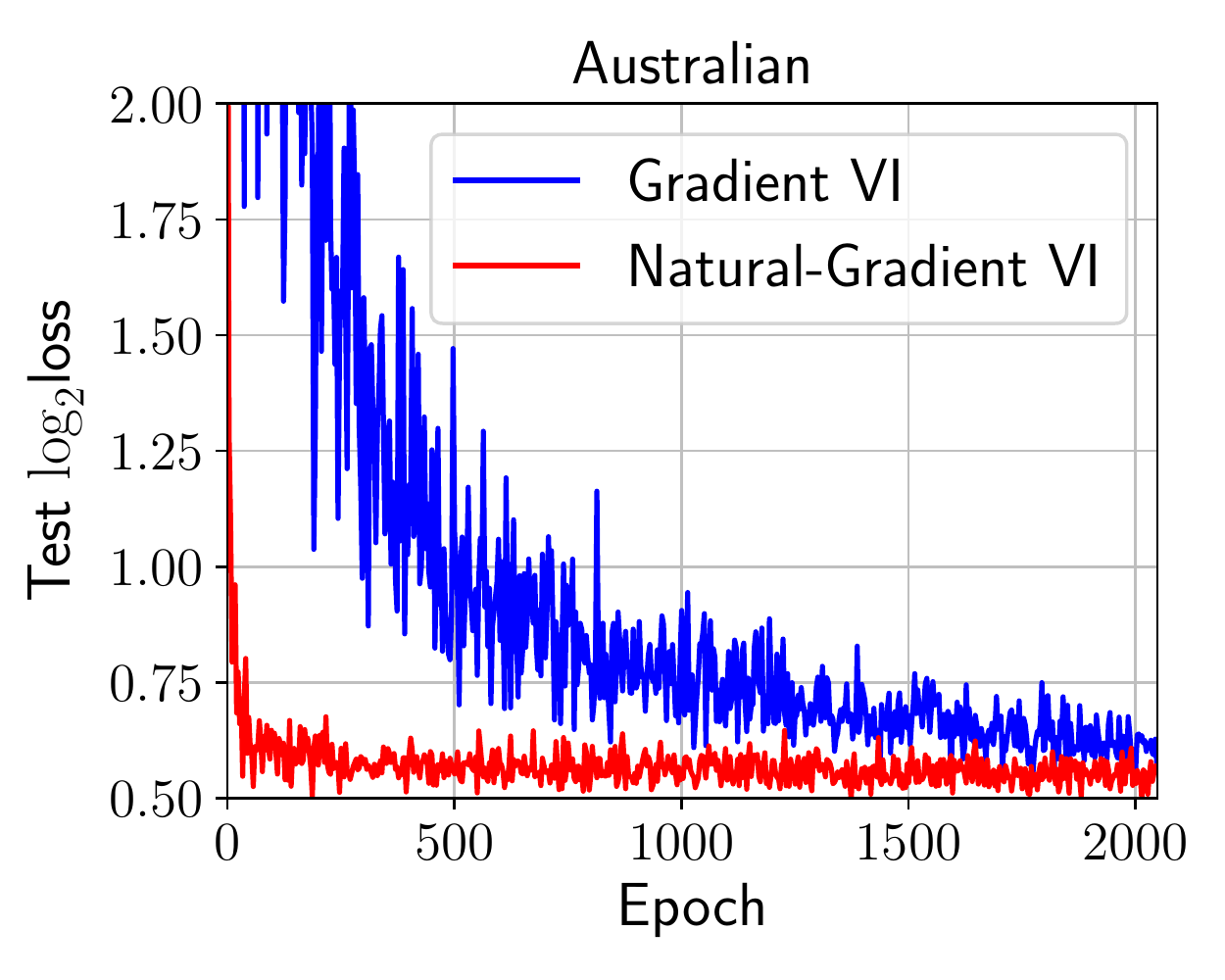}
             \includegraphics[height=1.9in]{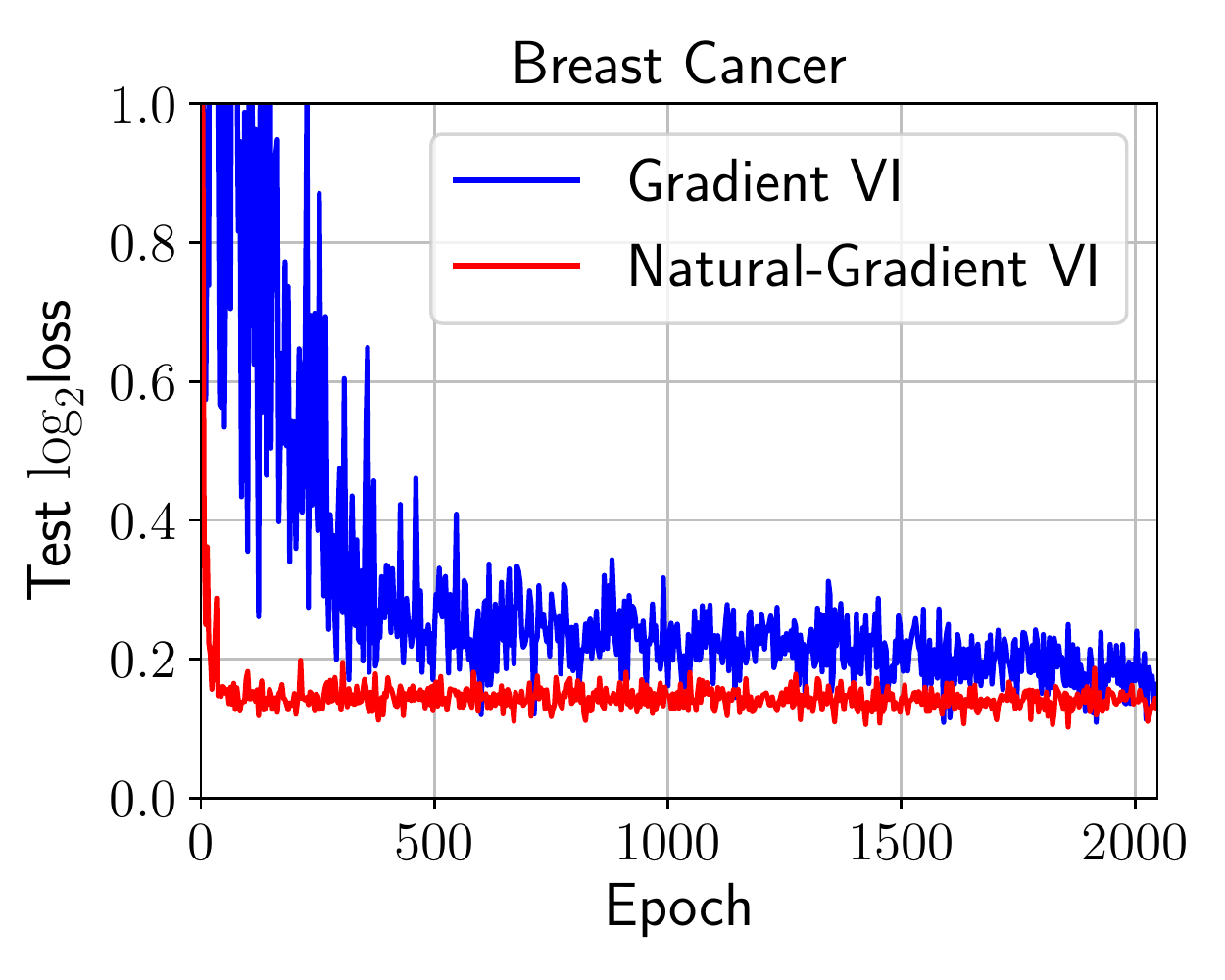} \label{figure:bnn}}
\caption{}
\label{figure:fig1}
\end{figure*}

An issue in using SGD is that it ignores the information geometry of the posterior approximation (see Figure \ref{figure:gaussians}).
Recent approaches address this issue by using stochastic \emph{natural-gradient} descent methods which exploit the Riemannian geometry of exponential-family approximations to improve the rate of convergence \cite{hoffman2013stochastic, honkela2004unsupervised, hensman2012fast}.
Unfortunately, these approaches only apply to a restricted class of models known as \emph{conditionally-conjugate} models, and do not work for nonconjugate models such as Bayesian neural networks.

This paper discusses some recent methods that generalize the use of natural gradients to such large and complex nonconjugate models. We show that, for exponential-family approximations, a duality between their natural and expectation parameter-spaces enables a simple natural-gradient update. The resulting updates are equivalent to a recently proposed method called Conjugate-computation Variational Inference (CVI) \cite{khan2017conjugate}.
An attractive feature of the method is that it naturally obtains \emph{local} exponential-family approximations for individual model components. We discuss the application of the CVI method to Bayesian neural networks and show some recent results from a recent work \cite{vadam2018} demonstrating faster convergence of natural-gradient VI methods compared to gradient-based VI methods (see Figure \ref{figure:bnn}).

\section{Problem Formulation}
In this section, we discuss the problem of variational inference and show how SGD can be used to optimize it. SGD ignores the geometry of the posterior approximations, and we discuss how natural-gradient methods address this issue. We end the section by mentioning issues with existing natural-gradient methods for variational inference.

	\subsection{Variational Inference (VI)}
	\label{sec:vi}
	We consider models\footnote{Methods discussed in this paper apply to a more general class of models, e.g., the model class discussed in \cite{khan2017conjugate}, but for clarity of presentation we focus on a restricted class.}
 that take the following form:
\begin{align}
   p(\data,\vlat) \propto \sqr{\prod_{i=1}^N p(\data_i|\vlat)} p(\vlat), \textrm{ where } \data := \{\data_i\}_{i=1}^N
   \label{eq:model}
\end{align}
where $p$ is a likelihood function which relates the model parameters $\vlat$ to the $i$'th data-example $\data_i$, and $p(\vlat)$ is the prior distribution which we assume to be an exponential-family distribution \cite{WainwrightJordan08},
\begin{align}
   p(\vlat) \propto h(\vlat)\exp\crl{\vphi(\vlat)^\top\vpriorpar - A(\vpriorpar)}, \textrm{ where } \vpriorpar\in\Omega
   \label{eq:prior}
\end{align}
where $\vphi$ is a vector of sufficient statistics, $\vpriorpar$ is the natural-parameter vector, and $A(\vnat)$ is the log-partition function.
The model parameter is a random vector here and sometimes is referred to as the \emph{latent} vector.

\begin{example}
   \emph{Consider Bayesian neural networks (BNN) \cite{bishop2006pattern} to model data $\data_i$ that contains input $\vx_i\in\real^D$ and a scalar output $y_i$. The vector $\vlat$ is the vector of network weights. The likelihood $p(\data_i|\vlat)$ could be an exponential-family distribution $p(y_i|f_{\bf{\lat}}(\vx_i))$ whose parameter $f_{\bf{\lat}}(\cdot)$ is a neural network parameterized by $\vlat$. We assume an isotropic Gaussian prior $p(\vz) :=
   \gauss(\vlat|0,\vI/\tau)$ where $\tau$ is a scalar. Its natural parameters are $\vpriorpar := \{0, -\tau\vI/2\}$.}
\qed
\end{example}

For such models, Bayesian approaches can estimate a measure of uncertainty by using the posterior distribution: $p(\vlat| \data) := p(\data|\vlat) p(\vlat)/p(\data)$.
This requires computation of the normalization constant $p(\data) = \int p(\data|\vlat) p(\vlat) d\vlat$ which unfortunately is difficult to compute in models such as Bayesian neural networks. One source of difficulty is that the likelihood $p(\data_i|\vlat)$ does not take the same form as the prior with respect to $\vlat$, or, in other words, the model is \emph{nonconjugate} \cite{gelman2014bayesian}. As a result, the product $p(\data|\vlat) p(\vlat)$ does not take a form with which $p(\data)$ can be easily computed.

Variational inference (VI) simplifies the problem by approximating $p(\vlat| \data)$ with a distribution $q(\vlat)$ whose normalizing constant is relatively easier to compute. In models \eqref{eq:model}, a straightforward choice is to choose $q(\vlat)$ to be of the same parametric\footnote{This restriction may not lead to a suboptimal approximation, e.g., in mean-field approximation in conjugate exponential-family models, the optimal form according to the variational objective turns out to be an exponential-family
   approximation
   \cite{bishop2006pattern}.} form as the prior $p(\vlat)$ but with a different natural-parameter vector $\vnat$, i.e., $q_\nat(\vlat) := h(\vlat) \exp[\vnat^\top \vphi(\vlat) - A(\vnat)]$. The parameter $\vnat$ can be obtained by maximizing the variational
objective which is also a lower bound to $p(\data)$
\cite{bishop2006pattern},
\begin{align}
  \max_{\nat\in\Omega} \elbofinal(\vnat) := \myexpect_{q_\nat} \sqr{ \log \frac{p(\vlat)}{q_\nat(\vlat)}} + \sum_{i=1}^N \myexpect_{q_\nat} [\log p(\data_i|\vlat)] . \label{eq:lb} 
\end{align}
where $\Omega$ is the set of valid variational parameters.
Intuitively, the first term favors $q_\nat(\vlat)$ which is close to the prior $p(\vlat)$ while the second term favors those that obtain high expected log-likelihood values. The variational objective has a very familiar form similar to many other \emph{regularized} optimization problems in machine learning \cite{bishop2006pattern}. 

\begin{example}
\emph{In the BNN example, we can choose $q_\nat(\vlat) = \gauss(\vlat|\vm,\vV)$ where $\vm$ is the mean and $\vV$ is the covariance. The natural-parameter vector is $\vnat:=\{\vV^{-1}\vm, -\half \vV^{-1}\}$, and our goal in VI is to maximize $\elbofinal$ with respect to these parameters.}
\qed
\end{example}

	\subsection{VI with Gradient Descent}
	\label{sec:vigrad}
	A straightforward approach to maximize $\elbofinal$ is to use a gradient-based method, e.g., the following stochastic-gradient descent (SGD) algorithm:
\begin{align}
   \vnat_{t+1} = \vnat_t + \rho_t \sqr{ \widehat{\nabla}_{\nat} \elbofinal (\vnat_t) } ,
   \label{eq:sgd}
\end{align}
where $t$ is the iteration number, $\rho_t$ is a step size, and $\widehat{\nabla}_{\nat} \elbofinal (\vnat_t)$ is a stochastic estimate of the derivative of $\elbofinal$ at $\vnat = \vnat_t$ (the `hat' here indicates a stochastic estimate).
Such stochastic gradients can be easily computed using methods such as REINFORCE \cite{williams1992simple} and the reparameterization trick \cite{kingma2013auto, rezende2014stochastic}.
This results in a simple but powerful approach which applies to many models and scales to large data.

Despite this, a direct application of SGD to optimize $\elbofinal(\vnat)$ is problematic because SGD ignores the \emph{information geometry} of the distribution $q_\nat(\vlat)$. 
To see this, we can rewrite \eqref{eq:sgd} as,
   \begin{align}
      \vnat_{t+1} = \arg\max_\lambda \,\, \vnat^\top \sqr{\widehat{\nabla}_{\nat} \elbofinal (\vnat_t) }  - \frac{1}{2\rho}\|\vnat - \vnat_t\|^2,
      \label{eq:sgd1}
   \end{align}
Equivalence can be established by taking the derivative and setting to 0.
The equation \eqref{eq:sgd1} implies that SGD moves in the direction of the gradient while remaining close, in terms of the Euclidean distance, to the previous $\vnat_t$. However, the Euclidean distance between natural parameters is not appropriate because $\vnat$ is the parameter of a distribution and the Euclidean distance is often a poor measure of
dissimilarity between distributions. This is illustrated in Figure \ref{figure:gaussians}. 
A more informative measure such as a Kullback-Leibler (KL) divergence, which directly measures the distance between distributions, might be more appropriate.

	\subsection{VI with Natural-Gradient Descent}
	\label{sec:vinatgrad}
	The issue discussed above can be addressed by using natural-gradient methods that exploit the information geometry of $q$ \cite{amari1998natural}. An exponential-family distribution induces a Riemannian manifold with a metric defined by the Fisher Information Matrix (FIM) \cite{amari2016information}, e.g. the FIM can be obtained as follows in the natural parameterization,
\begin{align}
   \fim(\vnat) := \myexpect_{q_\nat} \sqr{\nabla_\nat \log q_\nat(\vlat)\,\, \nabla_\nat \log q_\nat(\vlat)^\top}
\end{align}
Natural-gradient descent modifies the SGD step \eqref{eq:sgd1} by using the Riemannian metric instead of the Euclidean distance, 
   \begin{align}
      \max_\lambda \,\, \vnat^\top \sqr{\widehat{\nabla}_{\nat} \elbofinal (\vnat_t) }  - \frac{1}{2\alpha_t}(\vnat - \vnat_t)^\top \vF(\vlambda_t) (\vnat - \vnat_t),
   \end{align}
   where $\alpha_t>0$ is a scalar step size. 
   This results in an update similar to the SGD update shown in \eqref{eq:sgd},
   \begin{align}
      \vnat_{t+1} = \vnat_t + \alpha_t \sqr{\vF(\vnat_t) }^{-1}  \widehat{\nabla}_\nat\elbofinal (\vnat_t) ,
      \label{eq:ngd}
   \end{align}
   where the stochastic gradient is scaled by the FIM.
   The scaled stochastic-gradient is referred to as the stochastic \emph{natural gradient} defined as follows:
   \begin{align}
      \widetilde{\nabla}_\nat\elbofinal(\vnat) := \sqr{\vF(\vnat) }^{-1}  \widehat{\nabla}_\nat\elbofinal (\vnat) .
      \label{eq:nat_grad}
   \end{align}
   We use the notation $\tilde{\nabla}$ to differentiate the \emph{natural} gradient as opposed to the standard gradient in Euclidean space denoted by $\nabla$. In practice, the scaling, in a similar spirit to Newton's method, improves convergence and also simplifies step-size tuning.

   Natural gradients are also naturally suited for VI in certain class of models. A recent work in \cite{hoffman2013stochastic} shows that for conjugate exponential-family models, natural-gradients with respect to the natural-parameterization take a very simple form. For example, consider the first term in \eqref{eq:lb} which consists of the ratio of two terms that are conjugate to each other. The natural-gradient then is equal to the difference in the natural parameter of the two terms (see Eq. 41 in \cite{khan2017conjugate} for more details):
\begin{align}
   &\widetilde{\nabla}_\nat \myexpect_{q_\nat} \sqr{ \log \frac{p(\vlat)}{q_\nat(\vlat)}} \label{eq:nat_grad_conj} \\
   &= \sqr{\vF(\vnat) }^{-1} \widehat{\nabla}_\nat \myexpect_{q_\nat} \sqr{\vphi(\vlat)^T(\vpriorpar - \vnat) + A(\vnat)} = \vpriorpar - \vnat \nonumber
\end{align}
The above natural gradient does not require computation of the FIM, which is surprising. It is natural to ask whether a similar expression is possible when the model contains nonconjugate terms? We show that it is possible to do so if we perform natural-gradient descent in the natural parameter space, but not if we do it in the space of expectation parameters.

\section{Natural Gradients with Exponential Family}
\label{sec:natgradexp}
In this section, we show that natural gradient with respect to the natural parameters can be obtained by computing the \emph{gradient} with respect to the expectation parameter. In the next section, we will show that this enables a simple natural-gradient update which does not require an explicit inversion of FIM.

We start by defining the expectation\footnote{Sometimes also called the mean or moment parameter.} parameter $\vmean\in\real^M$ of an exponential-family distribution $q_\nat$ as follows: $\vmean(\vnat) := \myexpect_{q_\nat} \sqr{\vphi(\vz)}$, 
where we have expressed $\vmean$ as a function of $\vnat$.
Alternatively, $\vmean$ can be obtained from the natural parameters by simply differentiating the log-partition function, i.e., $\vmean(\vnat) = \nabla A(\vnat)$.
The mapping $\nabla A$ is one-to-one and onto (a bijection) iff the representation is minimal. Therefore, we can express $\elbofinal(\vnat)$ in terms of $\vmean$. We denote the new objective by $\elbofinal_*(\vmean) := \elbofinal(\vnat)$. We can now state our claim.

\begin{thm}
   For an exponential-family in the minimal representation, the natural gradient with respect to $\vnat$ is equal to the gradient with respect to $\vmean$, and vice versa, i.e.,
	\begin{align}
		\widetilde{\nabla}_\nat \elbofinal(\vnat) = \nabla_\mean \elbofinal_*(\vmean) \textrm{ and }
		\widetilde{\nabla}_\mean \elbofinal_*(\vmean) = \nabla_\nat \elbofinal(\vnat)
      \label{eq:dualgrad}
	\end{align}
\end{thm}
\begin{myproof}
\emph{
   Using chain rule, we can rewrite the derivative with respect to $\vnat$ in terms of $\vmean$: 
   \begin{align}
      \nabla_\nat \elbofinal (\vnat) &= \sqr{\nabla_\nat \vmean}  \nabla_\mean \elbofinal_*(\vmean) = \sqr{\nabla^2_{\nat\nat} A(\vnat)}  \nabla_\mean \elbofinal_*(\vmean) 
	\label{eq:duality}
   \end{align}
   It is well known that the second derivative of $A(\vnat)$ is equal to the FIM for exponential-family distribution, i.e., $\vF(\vnat) := \nabla_{\nat\nat}^2 A(\vnat)$ \cite{nielsen2009statistical}. This matrix is invertible when the representation is minimal. Therefore multiplying the above equation with inverse of $\vF(\vnat)$ gives us the first equality.
Since the FIM with respect to $\vnat$ is inverse of the FIM with respect to $\vmean$ \cite{nielsen2009statistical}, the second equality is immediate.}
\qed
\end{myproof}

This result is a consequence of a relationship between $\vmean$ and $\vnat$. The two vectors are related through the \emph{Legendre transform} which is the following transformation $\vmean = \nabla A(\vnat)$. Since $A(\vnat)$ is a convex function, the space of $\vmean$ and $\vnat$ are both Riemannian manifolds which are also duals\footnote{In information geometry, this is known as the \emph{dually-flat Riemannian structure} \cite{amari2016information}.} of each other. An attractive property of this structure is that the FIM in one space is the inverse of the FIM in the other space. This enables us to compute natural gradient in one space
using the gradient in the other, as shown in \eqref{eq:dualgrad}. 
This result is also discussed in an earlier work by Hensman et al. \cite{hensman2012fast} in the context of conjugate models, although they do not explicitly mention the connection to duality.

The natural gradient $\widetilde{\nabla}_\nat$ makes a better choice for conjugate models because $\nabla_\mean$ assumes a simple form which does not require computation of the FIM. The $\widetilde{\nabla}_\mean$ unfortunately does not have this property. For example, $\widetilde{\nabla}_\mean$ for \eqref{eq:nat_grad_conj} requires computation of the FIM because it is equal to $\vF(\vnat) (\vpriorpar - \vnat)$. This can be shown by using \eqref{eq:dualgrad}, \eqref{eq:nat_grad} and
\eqref{eq:nat_grad_conj}. 

The recent work by \cite{khan2017conjugate} propose to use the gradients with respect to $\vmean$ to perform natural gradient with respect $\vnat$. They arrive at this conclusion by using the equivalence of mirror descent and natural-gradient descent. Our discussion above complements their work by using the duality of the two spaces.

\section{Natural Gradients for Nonconjugate Models}
\label{sec:natgradnonconjugate}
In this section, we show that in some cases the natural gradient of the nonconjugate term can be easily computed by using $\nabla_\mean$. We also show that the resulting update takes a simple form.

We start with the expression for $\widetilde{\nabla}_\nat \elbofinal$. Using \eqref{eq:dualgrad} and \eqref{eq:nat_grad_conj}, it is straightforward to write this expression:
\begin{align}
   \widetilde{\nabla}_\nat\elbofinal(\vnat) := \vpriorpar - \vnat +  \sum_{i=1}^N \,\widehat{\nabla}_\mean \,\myexpect_q [\log p(\data_i|\vlat)] 
   \vert_{\mean=\mean(\nat)} ,
   \label{eq:natgrad_elbo}
\end{align}
where we have expressed $\vmean$ as a function of $\vnat$.
For notational convenience, we will denote $i$'th term inside the summation by  $\vgradmean_i(\vnat) := \widehat{\nabla}_\mean \myexpect_q [\log p(\data_i|\vlat)] \vert_{\mean=\mean(\nat)}$.

A \emph{stochastic natural-gradient descent update} can be obtained by using the gradient of a randomly sampled data example $\data_i$ and multiplying it by $N$, as shown below:
\begin{align}
   \vnat_{t+1} = (1- \alpha_t) \vnat_t + \alpha_t \sqr{\vpriorpar + N \vgradmean_i(\vnat_t)} ,
   \label{eq:cvi}
\end{align}
where the gradient is multiplied by $N$ to obtain an unbiased stochastic gradient. This update is equivalent to the update obtained in \cite{khan2017conjugate} where it is referred to as Conjugate-computation variational inference (CVI). In \cite{khan2017conjugate}, this is derived using a mirror-descent formulation, while we use the duality of the exponential family (Theorem 1).

 Unlike the SGD update, the natural-gradient update \eqref{eq:cvi} only computes gradients of the nonconjugate terms, thereby requires less computation. We now give an example which shows that $\vgradmean_i$ assumes a simple form and can be computed easily using automatic-gradient methods.
\\

\begin{example}
   \emph{For the BNNs example, $\vgradmean_i(\vnat)$ can be obtained by using backpropagated gradients $\vg_i(\vz) := \nabla_{\bf z} \log p(y_i|f_{\bf z}(\vx_i))$ and Hessians $\vH_i(\vz) := \nabla_{\bf zz}^2 \log p(y_i|f_{\bf z}(\vx_i))$. For a Gaussian $q_{\nat}$, there are two expectation parameters: $\vmean_1 := \myexpect_{q_\nat}(\vlat) = \vm$ and $\vmean_2 := \myexpect_{q_\nat}(\vlat\vlat^T) = \vm\vm^\top + \vV$, and two natural parameters: $\vnat_1 := \vV^{-1}\vm$ and $\vnat_2 := - \frac{1}{2}\vV^{-1}$. As shown in \cite{vadam2018}, we can write gradients as follows: 
\begin{align}
   \nabla_{\mean_1} \myexpect_{q_\nat} [\log p(y_i|f_{\bf z}(\vx_i))] &= \myexpect_{q_\nat} [\vg_i(\vlat)] - 2 \myexpect_{q_\nat} [\vH_i(\vlat)] \vm \nonumber\\
   \nabla_{\mean_2} \myexpect_{q_\nat} [\log p(y_i|f_{\bf z}(\vx_i))]&= \myexpect_{q_\nat} [\vH_i(\vlat)].
\end{align}
If we approximate the expectations using a single Monte Carlo sample $\vz_t \sim q_\nat(\vlat)$, we can write the update in \eqref{eq:cvi} as
\begin{align}
\label{eq:gaussupdates}
\vm_{t+1} &= \vm_t - \alpha_t \vV_{t+1} \left[\tau\vm_t - N\vg_i(\vlat_t) \right] \\
\label{eq:gaussupdates1}
\vV_{t+1}^{-1} &= (1-\alpha_t)\vV_t^{-1} + \alpha_t \left[\tau\vI - N\vH_i(\vlat_t) \right].
\end{align}
These updates take a form similar to Newton's method. The covariance matrix $\vV_t$ plays a similar role to the Hessian in Newton's method and scales the gradient in the update of $\vm_t$. The matrix itself contains a moving average of the past Hessians. 
It is, however, not common to compute Hessians for deep models, but, as we discuss in Section \ref{sec:application}, we can use another approximation to simplify this computation. With such an approximation, these updates can be implemented efficiently within existing deep learning code-bases as discussed in \cite{vadam2018}.
\qed
}
\end{example}

Similarly to the above example, it might be possible to employ automatic-gradient methods to compute natural gradients in many models. A recent work \cite{salimbeni2018natural} explores this possibility. Another stochastic approximation method discussed in \cite{salimans2013fixed} is also useful. For simple models, such as generalized linear models, where we can directly derive the distribution of the local variables, we can locally compute the gradients.
This is discussed in \cite{khan2017conjugate} for generalized linear models, Gaussian processes, and linear dynamical systems with nonlinear likelihoods.

\section{Local Approximations with Natural Gradients}
\label{sec:localapprox}
We now show that natural gradients not only result in simple updates, but they also give rise to local exponential-family approximations of the nonconjugate terms. An attractive feature of these approximation is that the natural gradient of a nonconjugate likelihood is also the natural parameter of its local approximation.

We start by analyzing the optimality condition of $\elbofinal$. First, by setting $\eqref{eq:natgrad_elbo}$ to zero, we note that a maximum $\vnat_*$ of $\elbofinal$ satisfies the following\footnote{We note that a similar optimality condition is used in \cite{salimans2013fixed} although the connection to natural gradients is not discussed.}:
$\vnat_* = \vpriorpar +  \sum_{i=1}^N  \vgradmean_i(\vnat_*)$.
Then, multiplying by $\vphi(\vz)$, exponentiating the whole equation, and by using the definition \eqref{eq:prior} of the prior, we can rewrite the optimality condition as follows,
\begin{align}
	q(\vlat|\vnat_*) \propto \sqr{\prod_{i=1}^N e^{\boldsymbol{\phi}(\mathbf{z})^\top {\bf \gradmean}_i(\boldsymbol{\nat_*})} } p(\vlat) 
   \label{eq:optnatgrad}
\end{align}
Comparing this update to the original model \eqref{eq:model}, we see that the nonconjugate likelihoods are replaced by \emph{local exponential-family approximations} whose natural parameters are the \emph{local natural-gradients} $\vgradmean_i(\vnat_*)$. This type of local approximation is employed in Expectation Propagation (EP) \cite{Minka01b}. In contrast, here they naturally emerge during a \emph{global} step, i.e., during the optimization of the whole variational objective.

We denote the $i$'th local approximation at iteration $t$ by $\tilde{q}_t^{(i)}$ and define it as follows,
\begin{align}
   p(\data_i|\vlat) \approx \tilde{q}_t^{(i)}(\vlat) \,\, \propto \,\, h(\vlat)\,\, e^{ \boldsymbol{\phi}(\boldsymbol{\lat})^\top \boldsymbol{\gradmean}_i(\boldsymbol{\nat}_t) },
\end{align}
We can then write the update \eqref{eq:cvi} as an \emph{approximate} Bayesian filter as shown below,
   \begin{align}
      q_{\nat_{t+1}}(\vlat) \propto  \Big[q_{\nat_t}(\vlat)\Big]^{1-\alpha_t} \Big[ \crl{\tilde{q}_t^{(i)}(\vlat)}^N p(\vlat) \Big]^{\alpha_t} .
   \end{align}
This update replaces each likelihood term in the model \eqref{eq:model} by the $i$'th likelihood term, which is why $\tilde{q}_t^{(i)}$ is raised to the power $N$.
All distributions in the above update take the same exponential form as $q_\nat$, and therefore the resulting computation can therefore be performed using \emph{conjugate computations}, i.e., by simply adding their natural parameters. This algorithm is referred to as Conjugate-computation VI (CVI) in \cite{khan2017conjugate}.
   
   Finally, if the parameters $\vpriorpar$ of the prior distribution do not change with iterations, then we can further simplify the updates by pulling $p(\vlat)$ out of the iterations and expressing the local natural-parameters, denoted by $\tilde{\vnat}^{(i)}$, as a recursion as shown below,
\begin{align}
   &q(\vlat|\vnat_{t+1}) \propto \sqr{\prod_{i=1}^N e^{\boldsymbol{\phi}(\mathbf{z})^\top \tilde{\boldsymbol{\nat}}_t^{(i)}} } p(\vlat) \\ 
   &\textrm{ where } \tilde{\vnat}_t^{(i)} = (1-\alpha_t) \tilde{\vnat}_{t-1}^{(i)} + \alpha_t \delta_{i,t} \sqr{N  \vgradmean_i(\vnat_t)},\,\, \forall i \nonumber
\end{align}
where $\delta_{i,t} = 1$ if $i$'th data point is selected in the $t$'th iteration. The natural-parameter $\tilde{\vnat}^{(i)}$ plays a similar role to the so-called \emph{site parameters} in EP \cite{Rasmussen06}. As the algorithm progresses, the local natural parameters converge to the optimal natural parameters $\vgradmean_i(\vnat_*)$ shown in \eqref{eq:optnatgrad}.

\section{Results on Bayesian Neural Networks}
\label{sec:application}
In this section, we compare an approximate natural-gradient VI method with a gradient-based VI method. The natural-gradient method employs two approximations to the update \eqref{eq:gaussupdates}-\eqref{eq:gaussupdates1}. The first approximation is to use a diagonal covariance matrix which enables a fast computation when dimensionality of $\vlat$ is large.
The second approximation is to use a \emph{generalized Gauss-Newton} approximation for the Hessian. This avoids the need to compute second-order derivatives making the implementation easier. The resulting method is called \emph{Variational Online Gauss-Newton (VOGN)} \cite{vadam2018}. The updates of this method, as discussed in \cite{vadam2018}, is very similar to the Adam optimizer \cite{kingma2014adam} and can be implemented with a few lines of code change. This makes it easy to apply VOGN to large deep-learning problems.

Figure \ref{figure:bnn} compares VOGN with a gradient-based approach called Bayes by Backprop \cite{blundell2015weight}. The latter optimizes $\elbofinal$ using the Adam optimizer.
The results are obtained using a neural network with single-hidden layer of 64 hidden units and ReLU activations. A prior precision of $\tau=1$, a minibatch size of 128 and 16 Monte-Carlo samples are used for all runs. The two figures show results on the following two datasets: `Australian' ($N=690$ and $D=14$) and `Breast Cancer' ($N=569$ and $D=10$) datasets. We show $\log_2$loss vs epochs, where a lower value indicates a better performance. 
We clearly see that the natural-gradient method is much faster than the gradient-based method.
See \cite{vadam2018} for more experimental results.

\section{Conclusions}
\label{sec:conclusions}
In this paper, we discuss methods for natural-gradient descent in variational inference. Unlike gradient-based approaches, natural-gradient methods exploit the information geometry of the solution and can converge quickly. We review a few recent works and provide new insights using the duality associated with exponential-family approximations. We discuss an attractive property of the natural-gradient to obtain local conjugate approximations for individual model components.
Finally, we showed some illustrative examples where these methods have been applied to perform Bayesian deep learning.

\section*{Acknowledgment}
\addcontentsline{toc}{section}{Acknowledgment}
We would like to thank the following people at RIKEN, AIP for discussions and feedback: Aaron Mishkin, Frederik Kunstner, Voot Tangkaratt, and Wu Lin. We would also like to thank James Hensman and Shun-ichi Amari for discussions.

\bibliographystyle{plain}
\bibliography{refs}

\end{document}